\documentclass[sigconf]{acmart}

\renewcommand\footnotetextcopyrightpermission[1]{} 
\usepackage[ruled,linesnumbered]{algorithm2e}

\SetCommentSty{mycommfont}

\usepackage{adjustbox}
\usepackage{xcolor}
\usepackage{subfigure}
\usepackage{graphicx}
\usepackage{booktabs}       
\colorlet{shadecolor}{gray!40}
\usepackage{tcolorbox}  
\usepackage{xspace}
\usepackage{enumitem}
\usepackage{amsmath}

\begin{document}

\title{Unlocking Cross-Lingual Sentiment Analysis through Emoji \\Interpretation: A Multimodal Generative AI Approach}

\author{Rafid Ishrak Jahan, Heng Fan, Haihua Chen, Yunhe Feng}
\affiliation{%
  \institution{University of North Texas, Denton, USA}
  \city{}
  \country{}}
\email{RafidIshrakJahan@my.unt.edu, Heng.Fan@unt.edu, Haihua.Chen@unt.edu, Yunhe.Feng@unt.edu}

\begin{abstract}

Emojis have become ubiquitous in online communication, serving as a universal medium to convey emotions and decorative elements. Their widespread use transcends language and cultural barriers, enhancing understanding and fostering more inclusive interactions. While existing work gained valuable insight into emojis understanding, exploring emojis' capability to serve as a universal sentiment indicator leveraging large language models (LLMs) has not been thoroughly examined. Our study aims to investigate the capacity of emojis to serve as reliable sentiment markers through LLMs across languages and cultures. We leveraged the multimodal capabilities of ChatGPT to explore the sentiments of various representations of emojis and evaluated how well emoji-conveyed sentiment aligned with text sentiment on a multi-lingual dataset collected from 32 countries. Our analysis reveals that the accuracy of LLM-based emoji-conveyed sentiment is 81.43\%, underscoring emojis' significant potential to serve as a universal sentiment marker. We also found a consistent trend that the accuracy of sentiment conveyed by emojis increased as the number of emojis grew in text. The results reinforce the potential of emojis to serve as global sentiment indicators, offering insight into fields such as cross-lingual and cross-cultural sentiment analysis on social media platforms. Code: \url{https://github.com/ResponsibleAILab/emoji-universal-sentiment}.

\end{abstract}


\maketitle

\section{Introduction and Background}
Emojis have evolved from simple novelties into powerful tools for expressing sentiments and emotions across diverse social media platforms. They serve as substitutes for non-verbal cues, such as facial expressions and gestures, which are often absent in text-based communication \cite{Bai_Dan_Mu_Yang_2019}. Due to their unique ability to enhance communication and enrich user experiences \cite{American_Psychological_Association}, emojis have become integral to digital interactions. Their universal appeal lies in their capacity to transcend linguistic and cultural barriers, making them a compelling subject of study, particularly in the domain of sentiment analysis. 

Recent research has provided valuable insight into understanding emoji sentiment utilizing emojis and their representations. For example, Novak et al.~\citep{kralj2015sentiment} engaged 83 human annotators to label the sentiment polarity (negative, neutral, or positive) of 751 emojis, employing the mean of the discrete probability distribution. Similarly, Gavilanes et al.~\citep{Fernandez-Gavilanes_Derks_et_al_2021} assessed the sentiment quality of emoji descriptions obtained from various online resources, including Emojipedia, Emojis Wiki, CLDR emoji character annotations, and iEmoji. Their analysis explored the extent to which these descriptions accurately reflect the intended sentiment of the emojis. However, these studies are constrained by their focus on European languages and the limited scope of emoji representations within their descriptions. Consequently, they fail to account for the influence of global linguistic diversity and diverse emoji representations—such as titles, pixel-level designs, and others—on capturing the intended sentiment of emojis.

Further analysis sheds light on predicting emoji usages based on text-based context. Zhao et al.~\citep{Zhao_Jia_An_Liang_Xie_Luo_2018} introduced a multitask multimodality gated recurrent unit to predict emoji usage in social media posts. Likewise, it was found by Qiu et al.~\citep{Qiu_Qiu_Lyu_Xiong_Luo_2024} that ChatGPT-4 (GPT-4) performs better than six LLMs when recommending emojis align with the semantics of a text. Yet, neither work portrays the picture of GPT-4 semantic preservation in the context of different languages and cultures. It is plausible that GPT-4 may occasionally generate inaccurate suggestions for languages impacted by low-resource language challenges \cite{yong2024low}.

Building on emoji-based research, other studies have highlighted the cosmopolitan nature of emojis and the potential of Generative AI understanding emojis. For example, Barbieri et al.~\citep{barbieri2016cosmopolitan} observed that the 150 most popular emojis preserve their semantics in four languages across four languages: American English, British English, Peninsular Spanish, and Italian. Additionally, Zhou et al.~\citep{Zhou_Xu_Wang_Lu_Gao_Ai_2024} reported that ChatGPT’s emoji annotations align closely with human interpretations. However, neither study explicitly examines the accuracy and consistency of emoji interpretations across a broader range of languages and cultural contexts.

While existing studies laid strong baselines for emoji research, gaps remain in addressing whether emojis have the potential to serve as a standalone medium of sentiment across languages. Moreover, traditional sentiment analysis often encounters challenges in cross-lingual contexts. Language-specific models are difficult to adapt to texts from various languages, and training multilingual models requires substantial computing resources \citep{10.1007/978-3-031-76806-4_8}. Additionally, being intricately tied to culture, sentiment does not always translate seamlessly across languages. It necessitates a universal medium that can serve as a bridge in multilingual sentiment analysis.

This paper investigates the feasibility of using emojis as a universal medium for sentiment indicators. Leveraging generative AI technologies, we analyze the sentiment of various emoji representations—including symbols, titles, descriptions, and pixel-level designs—across 19 languages and 32 countries. The contributions of this paper are summarized as follows:
\setlist[itemize,1]{left=0pt}
\begin{itemize}
    \item We investigated diverse multimodal representations of emojis for sentiment analysis using generative AI, demonstrating that emoji pixels, icons, and descriptions deliver the best performance.
    \item We proposed scalable, emoji-based, language-agnostic sentiment analysis approaches to infer the sentiment of text effectively.
    \item Our comprehensive cross-language evaluation highlights the potential of emojis as universal and standalone sentiment markers.
\end{itemize}

\section{Methodology}\label{sec:method}
This section begins by detailing the process of collecting various representations of emojis. Next, it describes the evaluation and selection of emoji representation combinations that achieve optimal performance for sentiment analysis. Finally, we introduce emoji-based approaches for sentiment analysis in textual data.

\subsection{Emoji Representation Collection}

Emojis can be represented in various formats, including icons, textual titles, descriptions, and pixel-based images, where icons refer to the visual representation of emojis, textual titles provide their concise names, descriptions offer detailed semantic explanations, and pixel-based images capture their graphical structure. To effectively evaluate the performance of generative AI models in sentiment analysis using different emoji representations, it is essential to explore individual and combined representations. The initial step towards this goal involves constructing a comprehensive emoji representation dataset encompassing all available emojis across these multiple formats. Given that no single data source offers a complete set of emoji representations covering all four formats (icons, titles, descriptions, and pixels), we developed an open-source tool\footnote{\url{https://github.com/ResponsibleAILab/emoji-universal-sentiment}\label{github}} to address this gap. Specifically, we aggregated and combined data from \href{https://emojipedia.org}{Emojipedia}\footnote{\url{https://emojipedia.org}} and the \href{https://www.unicode.org}{Unicode Consortium}\footnote{\url{https://www.unicode.org}} to construct our emoji representation dataset for 5030 emojis, ensuring a comprehensive and structured resource for further analysis.

\subsection{LLM-based Emoji Sentiment Estimation}

With diverse emoji representations established, we employ generative AI to estimate the sentiment of emojis across various representation formats. This sentiment can subsequently be used to infer the sentiment of multilingual text containing standalone emojis. Given that LLMs, such as ChatGPT, are trained on extensive datasets that inherently capture emoji sentiment, we propose that LLMs can provide a more optimal approach for estimating emoji sentiment compared to traditional manual labeling methods.

To determine the most suitable emoji representation for sentiment inference, we used Emoji Sentiment Ranking v1.0 (ESR v1.0)~\cite{emoji_sentiment_ranking_v1.0}, which offers a sentiment overview of 751 emojis based on category (positive, neutral, negative), as a benchmark for evaluation. The ESR v1.0 does not classify emoji as a single sentiment, instead, it calculates sentiment scores of each emoji category~\cite{emoji_sentiment_ranking_v1.0}. For example, an emoji can be predominantly positive but can also be used in neutral and negative contexts. Nonetheless, for our study, we evaluate the performance of each emoji against the predominant category of ESR v1.0. We first assessed the performance of individual emoji representations (icons, titles, descriptions, and pixels) and their various combinations, resulting in a total of $C_4^1 + C_4^2 + C_4^3 + C_4^4 = 15$ combinations. We employed ChatGPT-4o (GPT-4o) to conduct sentiment analysis using the following prompt.

{\small
\begin{tcolorbox}[colback=gray!5, colframe=black, boxrule=1pt, rounded corners, title=\textbf{GPT-4o prompt to classify emoji sentiment by combining emoji pixel, icon, and description representation}, fonttitle=\bfseries]
``Classify the sentiment of the following Emoji's picture by combining emoji icon and emoji description, and give one word answer from positive or negative or neutral.''
\end{tcolorbox}
}

Through our comprehensive evaluation of emoji representation combinations, as reported in Table \ref{tab:feature_intersections}, we found that the combination of emoji icons and textual descriptions achieved the best performance among pure text-based representations, while the combination of pixel, icon, and description outperformed all other multimodal representation combinations. The substantial performance gap highlights the importance of incorporating multimodality when analyzing emoji sentiments. Consequently, we adopted the best-performing representation to establish the ground truth for our dataset of 5,030 emojis. The resulting emoji sentiment dataset, built using GPT-4, has been made publicly available on GitHub$^1$ to ensure reproducibility and support further research.

\begin{table}[htbp]
    \centering
    \setlength{\tabcolsep}{9pt} 
    \renewcommand{\arraystretch}{1} 
    \caption{Performance of sentiment analysis with emoji representations by GPT-4o vs. ESR v1.0 (615 emojis)}
    \label{tab:feature_intersections}
    \begin{adjustbox}{width=\columnwidth,center}
    \begin{tabular}{lc}
        \toprule
        \textbf{Emoji Representation} & \textbf{Emojis Matched with ESR v1.0} \\
        \midrule
        Icon & 338 \\
        Title & \underline{303} \\
        Description & 357 \\
        Pixel & 332 \\
        Icon \& Description & 363 \\
        Icon \& Title & 349 \\
        Icon \& Pixel & 347 \\
        Title \& Description & 329 \\
        Title \& Pixel & 342 \\
        Pixel \& Description & 371 \\
        Icon \& Title \& Description & 348 \\
        Pixel \& Icon \& Title & 360 \\
        Pixel \& Icon \& Description & \textbf{373} \\
        Pixel \& Title \& Description & 365 \\
        Pixel \& Icon \& Title \& Description & 368 \\
        \bottomrule
    \end{tabular}
    \end{adjustbox}
\end{table}

\begin{figure}[htbp]
    \centering
    \includegraphics[width=\linewidth]{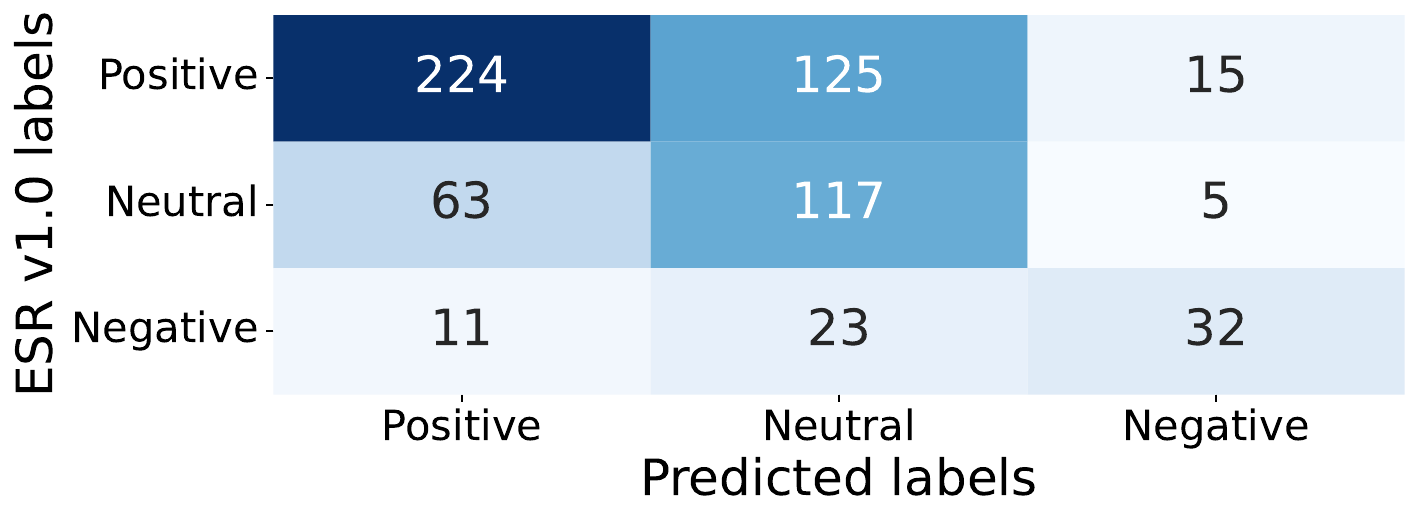}
    \vspace{-0.6cm}
    \caption{Sentiment confusion matrix regarding the best emoji representation and ESR v1.0}
    \vspace{-0.6cm}
    \label{fig:confusion-matrix}
\end{figure}

The performance of the best classification approach was evaluated using the $F_1$ score. Our approach involved assigning each emoji a single sentiment label. In contrast, the reference values for evaluation were derived from the ESR v1.0 dataset~\cite{emoji_sentiment_ranking_v1.0}, where each emoji's sentiment was determined by the highest sentiment category score. The $F_1$ scores were calculated based on the confusion matrix presented in Figure \ref{fig:confusion-matrix}, using the Pixel + Icon + Description emoji representation. For positive emoji classification, the $F_1$ score was 0.68, demonstrating that GPT-4o can reasonably identify positive emojis. For neutral and negative emoji classifications, $F_1$ scores of 0.52 and 0.54, respectively, indicate moderate performance. The relatively lower $F_1$ scores can be attributed to differences in emoji representations and methodologies between ESR and our approach.

\subsection{Standalone Emoji Sentiment Algorithms}\label{sec:standaloneemoji}

Since emojis maintain the same visual representation across different languages, they can serve as universal sentiment indicators for multilingual sentiment analysis. When a text contains only a single emoji, we use the sentiment conveyed by that emoji to represent the sentiment of the entire text. However, when text contains multiple emojis, we propose novel approaches to effectively combine the sentiments of multiple emojis to derive an overall sentiment.

Suppose a given text $t$ contains $n$ emojis $e = [e_1, e_2, \dots, e_n]$, with corresponding sentiments $s = [s_1, s_2, \dots, s_n]$, where $s_i \in \{\text{pos}, \text{neu}, \text{neg}\}$. To facilitate the calculation of $s_i$, we assign sentiment weights $w_{\text{pos}}, w_{\text{neu}}, w_{\text{neg}}$ to positive, neutral, and negative sentiments, respectively, where $w_{\text{pos}} > w_{\text{neu}} > w_{\text{neg}}$. Thus, the overall sentiment of the text $t$ can be represented as $\sum w_{s_i}$. The sentiment of $t$ is classified as follows: if $\sum w_{s_i} > \theta$, then the sentiment of $t$ is \textbf{positive}; if $\sum w_{s_i} = \theta$, then the sentiment of $t$ is \textbf{neutral}; and if $\sum w_{s_i} < \theta$, then the sentiment of $t$ is \textbf{negative}, where $\theta$ is a threshold to distinguish the positive and negative sentiments.

We evaluated three scoring systems for the emoji sentiment aggregation algorithm. The first scoring system, referred to as Basic Sentiment Aggregation (BSA), is represented as \(w_{\text{pos}} = 1, w_{\text{neu}} = 0, w_{\text{neg}} = -1\) for positive, neutral, and negative weights and $\theta = 0$. For the second scoring system, termed Dual Positive Model (DPM), we address mitigating the reduction of sentiment suppression since referring to neutral as $\theta$ makes them invisible in sentiment calculation. We considered neutral emojis as mild positives that act as a softener within the text. DPM is represented as \(w_{\text{pos}} = 2, w_{\text{neu}} = 1, w_{\text{neg}} = -2\) for positive, neutral, and negative weights and $\theta = 0$.

Lastly, we modified the Boyer-Moore majority vote algorithm where the final sentiment of emojis is determined by the sentiment category with the highest count of emojis. Let \( c_{\text{pos}} \), \( c_{\text{neu}} \), and \( c_{\text{neg}} \) represent the counts of positive, neutral, and negative emojis in a given text, respectively. 
    The sentiment is then determined as:
\[
sentiment = 
\begin{cases} 
\text{positive} & \text{if } c_{\text{pos}} > \max(c_{\text{neu}}, c_{\text{neg}}), \\
\text{neutral} & \text{if } c_{\text{neu}} > \max(c_{\text{pos}}, c_{\text{neg}}), \\
\text{negative} & \text{if } c_{\text{neg}} > \max(c_{\text{pos}}, c_{\text{neu}}).
\end{cases}
\]

\subsection{Emoji Position-Aware Sentiment Algorithm}

The proposed BSA, DPM, and majority voting algorithms described in Section~\ref{sec:standaloneemoji} do not account for the positions of emojis when inferring sentiment. To address this limitation, we further fine-tuned our BSA algorithm to capture subtle variations in sentiment based on emoji placement and repetition. BSA was selected for fine-tuning because it outperformed both the DPM and majority voting methods (see Table~\ref{tab:algorithmCompare}). The fine-tuning process introduced a nuanced weighting mechanism that reflects how users naturally convey sentiments with emojis. Specifically, the algorithm can prioritize the first and last emojis to capture introductory and concluding tones while emphasizing repeated and consecutive emoji patterns to highlight strong and consistent sentiments.

Following BSA, we set \(w_{\text{pos}} = 1, w_{\text{neu}} = 0, w_{\text{neg}} = -1\) for positive, neutral, and negative weights and $\theta = 0$. 
Sentiments are assigned based on the following criteria: 
\textbf{(1) Prioritize the first emoji (First)}: ${sentiment}_f = w_{s_1}$.
\textbf{(2) Prioritize consecutive emojis (Consec.)}: suppose $m$ unique emojis $[e_{c_1}, e_{c_2}, ..., e_{c_m}]$ are observed in $e=[e_1, e_2, ..., e_n]$ consecutively with consecutive length of $l=[l_1, l_2, ..., l_m]$, then ${sentiment}_c = \sum_{i=1}^{m} w_{s_{c_i}} \cdot l_i$, where $w_{s_{c_i}}$ represents the weight assigned to the sentiment of each unique emoji $e_{c_i}$. \textbf{(3) Prioritize repeated emojis (Repeat)}: suppose $m$ unique emojis $[e_{r_1}, e_{r_2}, ..., e_{r_m}]$ are observed in $e=[e_1, e_2, ..., e_n]$ with the repeated frequency of $f=[f_1, f_2, ..., f_m]$, then ${sentiment}_r = \sum_{i=1}^{m} w_{s_{r_i}} \cdot f_i$, where $w_{s_{r_i}}$ represents the weight assigned to the sentiment of each unique emoji $e_{r_i}$. \textbf{(4) Prioritize the last emoji (Last)}: ${sentiment}_l = w_{s_n}$. 
Finally, \textbf{(5) Aggregate all (All)}: ${sentiment}_{all} ={sentiment}_f + {sentiment}_c + {sentiment}_r +{sentiment}_l$.

\section{Experiment Results}\label{sec:experimentResults}
This section introduces the multilingual datasets used for emoji-based sentiment evaluation, outlines the sentiment ground truth establishment, and presents the results assessing emoji performance as a global sentiment marker.

\subsection{Multilingual Dataset for Sentiment Analysis}
To evaluate the performance of our methodologies of identifying text sentiment using standalone emojis in a multilingual context, we collected tweets from X (known as Twitter) related to the 2018 FIFA Soccer World Cup. Specifically, during the 2018 FIFA Soccer World Cup, we leveraged Twitter Streaming APIs to collect 32 sub-datasets from 32 participant countries and regions using geographic filters and World Cup-related keywords in their native languages. The entire sub-datasets cover 19 languages. For each participant country, we randomly selected 80,000 tweets, and the ground truth of sentiment was set up by GPT-4o, as it provides more 94\% accuracy on the long form of sentiment reviews \cite{belal2023leveraging}. The following prompt was used to analyze the sentiment expressed in tweets.
{\small
\begin{tcolorbox}[colback=gray!5, colframe=black, boxrule=1pt, rounded corners, title=\textbf{GPT-4o prompt to classify tweet sentiment as ground truth}, fonttitle=\bfseries]
``Find the sentiment of the following tweet by considering everything including the text, emoji, and URLs in the tweet, and give a one-word answer from `positive', `negative', and `neutral'. Please consider it as a casual tweet in which users express themselves. Language in a tweet can be informal and may not follow proper grammatical structure.''
\end{tcolorbox}
}

We also prompted GPT-4o to query whether it processes all 19 languages with equal accuracy. GPT-4o acknowledged varying accuracy levels, categorizing them into three distinct groups: high accuracy, good accuracy, and varied performance. To enhance consistency and accuracy, we conducted a second round of sentiment ground truth generation with GPT-4o, translating each text into English using the Google Translate API. Consequently, our datasets comprise original multilingual tweets from 32 countries with sentiment labels, along with their English translations and corresponding sentiment labels.

\subsection{Emoji Standalone Sentiment Analysis}

We investigated the feasibility of using emojis as standalone universal sentiment indicators across languages, utilizing tweet datasets spanning 19 languages and 32 countries. Table \ref{tab:algorithmCompare} presents the average sentiment accuracy for tweets from 32 countries when analyzed using only emojis versus text with emojis. We evaluated the performance of three algorithms—BSA, DPM, and Majority Voting—on both the original tweet data and their English translations. The results demonstrated that BSA achieved the highest accuracy of 77.71\%. Our experiments with translated text yielded results closely aligned with those of the initial experiment, reinforcing the robustness of the findings.

\begin{table}[htbp]
    \centering
    \setlength{\tabcolsep}{8pt} 
    \renewcommand{\arraystretch}{1}
    \caption{Average accuracy of sentiment analysis using BSA, DPM, and Majority Voting algorithms across 19 languages}
    \vspace{-0.4cm}
    \label{tab:algorithmCompare}
    \begin{adjustbox}{width=\columnwidth,center}
    \begin{tabular}{lccc}
        \toprule
        \textbf{Dataset} & \textbf{BSA} & \textbf{DPM} & \textbf{Majority Voting} \\
        \midrule
        Original Text & \textbf{77.71\%} & \underline{73.07\%} & 75.69\% \\
        Translated Text & \textbf{76.05\%} & \underline{71.02\%} & 74.15\% \\
        \bottomrule
    \end{tabular}
    \end{adjustbox}
\end{table}

We further evaluated the performance of emoji position-aware sentiment analysis algorithms, including prioritizing the first, consecutive, repeated, and last emojis, as well as a method aggregating all these strategies, under the same settings as BSA, as shown in Table \ref{tab:algorithmOptimize}. Prioritizing the first emoji in the sequence increased BSA accuracy to 81.43\%, indicating that the sentiment of the first emoji is closely aligned with the overall sentiment of the text.

\begin{table}[htbp]
    \centering
    \setlength{\tabcolsep}{8pt} 
    \renewcommand{\arraystretch}{1}
    \caption{Average accuracy of sentiment analysis by algorithms prioritizing first, consecutive, repeated, last, and aggregated emojis across 19 languages}
    \label{tab:algorithmOptimize}
    \vspace{-0.2cm}
    \begin{adjustbox}{width=\columnwidth,center}
    \begin{tabular}{lccccc}
        \toprule
        \textbf{Dataset} & \textbf{First} & \textbf{Consec.} & \textbf{Repeat} & \textbf{Last} & \textbf{All}\\
        \midrule
        Original Text & \textbf{81.43\%} & 80.42\% & 80.41\% & \underline{79.97\%} & 80.53\%  \\
        Translated Text & \textbf{79.45\%} & 77.52\% & 78.35\% & 77.98\% & \underline{77.45\%}  \\
        \bottomrule
    \end{tabular}
    \end{adjustbox}
\end{table}

\begin{figure}[htbp]
    \centering
    \includegraphics[width=\linewidth]{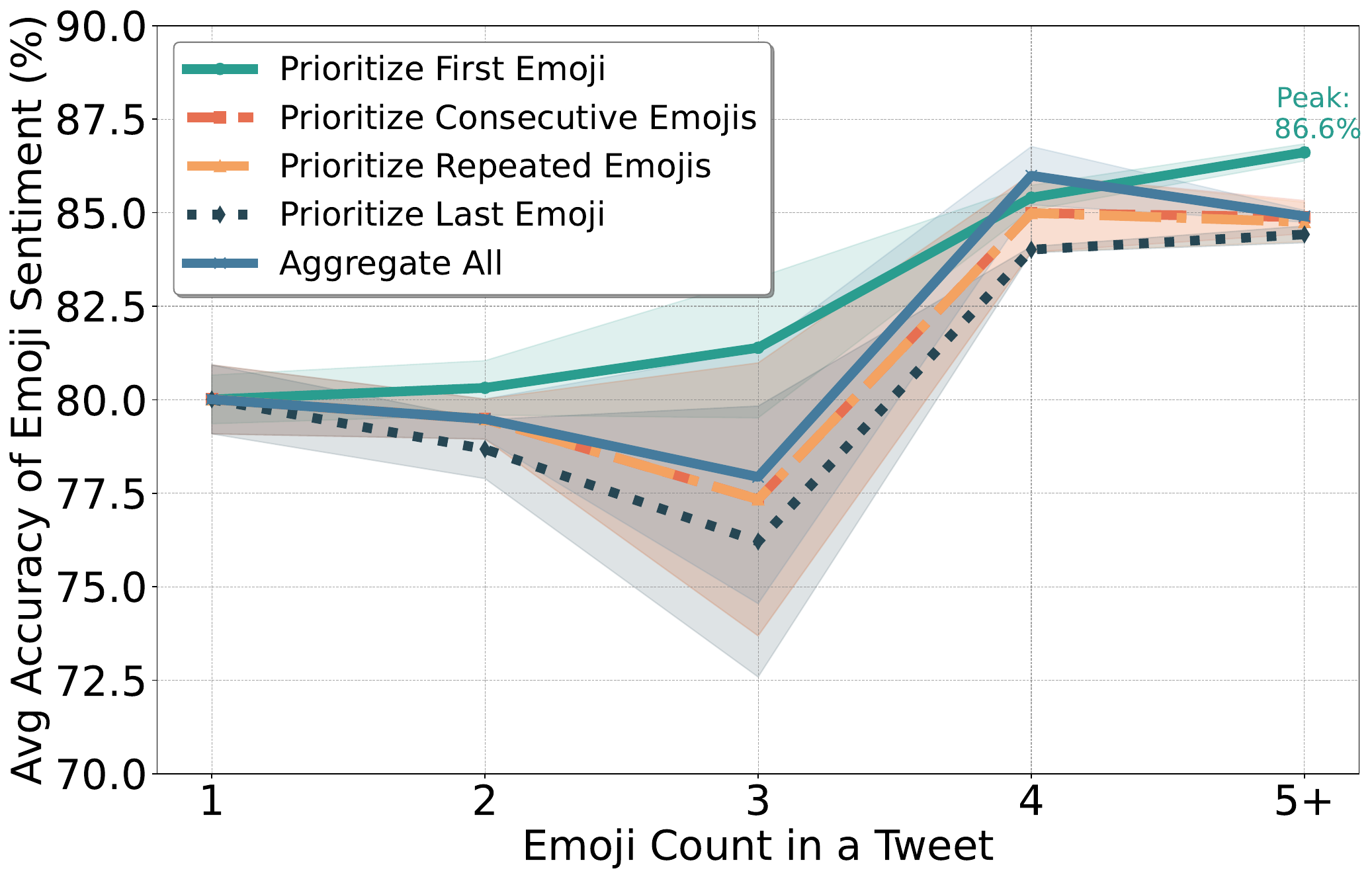}
    \vspace{-0.6cm}
    \caption{Sentiment accuracy for standalone emojis vs. text with emojis by emoji count}
    \vspace{-0.44cm}
    \label{fig:avgEmojiCount}
\end{figure}

Figure \ref{fig:avgEmojiCount} illustrates how the performance of algorithms prioritizing the first, consecutive, repeated, and last emojis, as well as a method aggregating these strategies, varies with the increasing number of emojis in a tweet. A gradual positive trend is observed between the number of emojis per tweet and sentiment accuracy, indicating a positive correlation: as the number of emojis increases, the accuracy of sentiment matching between emojis and text with emojis improves. However, Figure \ref{fig:avgEmojiCount} also reveals a performance dip in the ``2 to 3 emojis'' range for all algorithms except the one prioritizing the first emoji. Overall, our experimental results demonstrate that prioritizing the first emoji achieves the highest performance, with an accuracy of 81.43\% in sentiment analysis using emojis alone.

\section{Conclusion}\label{sec:conclude}
This study highlights the potential of emojis as universal sentiment indicators across languages and cultures, achieving an accuracy of 81.43\% on a multilingual dataset spanning 32 countries and 19 languages. By leveraging diverse emoji representations—such as pixel, icon, and description—our findings underscore the expressive power of non-textual elements like emojis in sentiment analysis. Notably, prioritizing the sentiment of the first emoji in a sequence closely aligns with the overall sentiment of a tweet, and combining multiple representations significantly outperforms single modalities. Furthermore, we observed a positive correlation between the number of emojis in a tweet and the accuracy of sentiment analysis for the algorithm that prioritizes the first emoji. These results emphasize the role of emojis in bridging linguistic and cultural gaps, offering valuable insights for future applications in cross-lingual natural language processing and social media analytics.

\bibliographystyle{ACM-Reference-Format}
\bibliography{custom}

\end{document}